\documentclass[runningheads]{llncs}
\renewcommand{\lastandname}{\unskip,}

\usepackage{accv}

\usepackage{accvabbrv}

\usepackage{graphicx}
\usepackage{booktabs}
\usepackage{caption}

\usepackage[accsupp]{axessibility}  

\usepackage[pagebackref,breaklinks,colorlinks,citecolor=accvblue]{hyperref}
\usepackage{hyperref}
\usepackage{multirow}
\usepackage{comment}
\usepackage{adjustbox}
\usepackage{float}
\usepackage{xcolor,colortbl}
\usepackage{xcolor}

\usepackage{orcidlink}

\begin{document}

\title{Frequency-Decomposed Avatar Representation for Varying Camera Distances}

\titlerunning{CloseUpAvatar}

\author{David Svitov\inst{1,2}\orcidlink{0009-0009-9116-0416} \and
Pietro Morerio\inst{2}\orcidlink{0000-0001-5259-1496} \and
Lourdes Agapito\inst{3}\orcidlink{0000-0002-6947-1092} \and
Alessio Del Bue \inst{2}\orcidlink{0000-0002-2262-4872}}

\authorrunning{D.~Svitov et al.}

\institute{Università degli Studi di Genova, Genoa, Italy \and
Istituto Italiano di Tecnologia (IIT), Genoa, Italy \and
Department of Computer Science, University College London\\
\email{\{david.svitov, pietro.morerio, alessio.delbue\}@iit.it \\ l.agapito@cs.ucl.ac.uk}}

\maketitle

\begin{abstract}
  We present a CloseUpAvatar - a novel approach for articulated human avatar representation supporting a wider range of camera motions, while preserving rendering quality for close-up views. CloseUpAvatar represents an avatar as a set of textured planes with frequency-decomposed learnable textures for low and high-frequency detail. The method automatically switches to high-frequency textures when the camera comes close to the avatar's surface and gradually reduces their impact as the camera moves farther away. Such parametrization of the avatar enables CloseUpAvatar to adjust rendering quality based on camera distance ensuring realistic rendering across a wider range of camera orientations than previous approaches. We provide experiments on the ActorsHQ dataset with high-resolution input images and the THuman4.0 dataset with diverse articulated poses. CloseUpAvatar demonstrates both qualitative and quantitative improvements over existing methods in rendering from novel wide range camera positions, while maintaining high FPS by limiting the number of required primitives. Project page: \\ \href{https://david-svitov.github.io/CloseUpAvatar_project_page}{https://david-svitov.github.io/CloseUpAvatar\_project\_page}
\end{abstract}

\section{Introduction}
\label{sec:intro}

\begin{figure}[t]
  \centering
  \includegraphics[width=12.0cm]{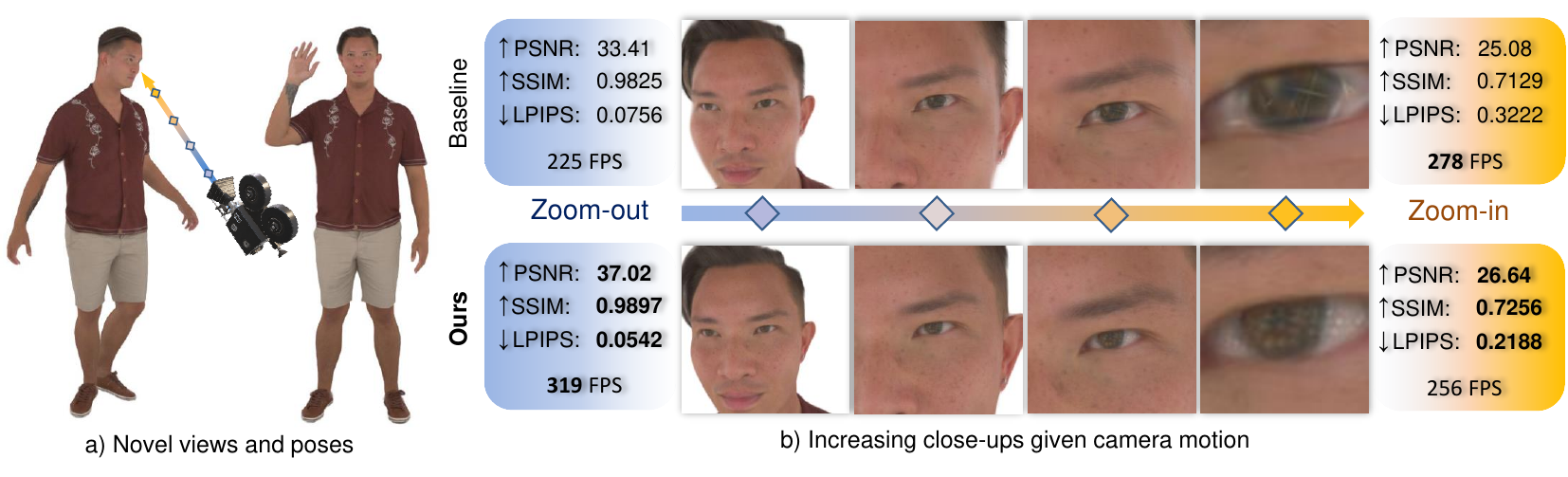}
   \caption{\textbf{High-quality rendering from different distances.} We present the method for (a) photo-realistic animatable avatars, (b) that is able to provide high-fidelity rendering under changing camera distance. In contrast with existing approaches, such as Mmlphuman~\cite{zhan2025real}, our method successfully avoids artifacts, maintaining rendering quality, and ensuring real-time rendering during extreme close-ups.}
   \label{fig:teaser}
\end{figure}

Digital human avatars are of paramount importance for such fields as telepresence, computer games, human-robot interaction and virtual reality (VR). Articulated full-body avatars indeed are more common for VR applications, as they enable a wide range of interaction scenarios. The user in VR spaces should be able to freely walk around the human avatar and interact with it close-up if needed. Unfortunately, existing methods for avatar generation either focus on highly realistic head avatars \cite{ma2021pixel, saito2024relightable, li2025tega, aneja2025scaffoldavatar} or on articulated full-body avatars \cite{Qian20233DGSAvatarAA, hu2024gaussianavatar, li2024animatable, zhan2025real}, which tend to reveal artifacts when looked at closely (\cref{fig:teaser}(b)). This work aims to develop a universal avatar representation that provides high-quality rendering at both close and far view distances.

Many existing works focused on realistic avatar generation \cite{Qian20233DGSAvatarAA, hu2024gaussianavatar, li2024animatable, zhan2025real, chen2024meshavatar, saito2019pifu, alldieck2022photorealistic, huang2020arch, he2021arch++, svitov2023dinar, Qian20233DGSAvatarAA, lei2024gart, hu2024gaussianavatar, svitov2024haha, bashirov2024morf, jiang2023instantavatar, weng2022humannerf} by utilizing different representations. Early approaches \cite{alldieck2019learning, alldieck2018detailed, wang2022arah, chen2021animatable} used a mesh or a Neural Radiance Field (NeRF)~\cite{mildenhall2021nerf} representation to render an avatar, though these methods had serious limitations: meshes lacked detail, while NeRF suffered from slow inference speed due to multiple MLP runs. Most recent approaches \cite{Qian20233DGSAvatarAA, hu2024gaussianavatar, li2024animatable, zhan2025real} utilize 3D Gaussian Splatting (3DGS) \cite{kerbl20233d} for avatar representation to tackle the challenge of real-time realistic rendering, but struggle to generalize across camera distances. 
A desired property is therefore consistent avatar rendering across camera distances, enabling interactions without a fixed viewing distance.

The task of adjusting detail levels for 3D scenes is known as Level of Detail (LOD)
\cite{luebke2002level} and focuses on selecting an appropriate number of details for each camera distance. Such methods often involve dynamic mesh topology adjusting \cite{william1992decimation, cohen1996simplification, garland1997surface} or reducing the number of primitives \cite{Ren2024OctreeGSTC, Yang2025Virtualized3G, Seo2024FLoDIF} to improve inference speed. Several avatar generation methods \cite{sun2024crowdsplat, dongye2024lodavatar} have successfully managed to increase rendering speed in crowded scenes by reducing detail for far views. CrowdSplat~\cite{sun2024crowdsplat} dynamically adjusts the number of Gaussians, but concentrates only on far-distanced cameras. LoDAvatar~\cite{dongye2024lodavatar} selectively changes the number of Gaussians based on the camera distance, but lacks of rendering realism due to the restriction of Gaussian positions with the underlying mesh. To the best of our knowledge, none of the existing avatar representations are explicitly designed to maintain both efficiency at far distances and fidelity in close-up views.

In this work, we propose separating low and high-frequency details across different sets of textures and using their weighted mixture to achieve high-quality rendering at both close and far views while preserving FPS. First, we developed a human avatar approach based on a textured 2DGS~\cite{huang20242d} representation that stores color information in corresponding textures for each Gaussian. Namely, we utilized BBSplat~\cite{svitov2024billboard} textured primitives that assign color and transparency textures to each 2DGS surfel in the scene. Such representation enables high fidelity of the avatars while maintaining a limited number of primitives with rich texture information.

We found that varying camera distance during training results in blurred artifacts in both 3DGS and textured surfels representations. We propose to circumvent this issue for textured primitives by using a frequency-decomposed avatar appearance representation to increase detail when necessary (\eg when we have close-ups). For each surfel, we create two sets of textures at different scales to represent low and high-frequency details. Namely, low-frequency color changes by the first set of small textures, and sharp details by the second set of larger textures, while the general color is represented by Spherical Harmonics (SH). We gradually interpolate the second set of textures only for close views based on the surfels' screen projection size. We ensure convergence of the method by developing a specialized training policy for frequency-decomposed textures.

To summarise, our contributions are as follows:
\begin{itemize}
    \item We proposed a novel frequency-decomposed avatar appearance representation for highly detailed rendering of full-body avatars, which for the first time enables high fidelity for close-up views while maintaining efficiency for distanced views;
    \item We developed a textured surfels-based approach to generate a full-body avatar from multi-view data and demonstrated the textured surfels' efficiency for animatable human representations;
    \item We experimentally demonstrate, using open data, that the proposed approach benefits when training with images taken from varying camera distances, while other methods struggle to converge with such data augmentation. 
\end{itemize}
 

\section{Related Works}
\label{sec:related_works}

\textbf{Full-body human avatars.} The avatar generation approaches differ by the type of input used: one-shot \cite{saito2019pifu, alldieck2022photorealistic, huang2020arch, he2021arch++, svitov2023dinar}, monocular video \cite{Qian20233DGSAvatarAA, lei2024gart, hu2024gaussianavatar, svitov2024haha, bashirov2024morf, jiang2023instantavatar, weng2022humannerf, raj2021anr, moon2024expressive}, and multi-view \cite{Qian20233DGSAvatarAA, hu2024gaussianavatar, li2024animatable, zhan2025real, chen2024meshavatar, wang2025relightable, li2022tava}. While one-shot and monocular video methods provide quick generation of avatars, the most realistic results are achieved with multi-view input data due to the extensive input information. To represent avatars, different approaches utilize different rendering techniques, including 3D meshes, NeRF~\cite{mildenhall2021nerf}, and 3DGS~\cite{kerbl20233d}. While 3D meshes were mostly used by early methods \cite{alldieck2019learning, alldieck2018detailed, alldieck2018video}, they still find application in the recent works \cite{svitov2024haha, chen2024meshavatar} but pose a challenge on realistic rendering. NeRF implicit representation is utilized by some works \cite{weng2022humannerf, wang2022arah, chen2021animatable} and enables photo-realistic rendering even with changing camera distance in static scenes \cite{barron2021mip}, but lacks training and rendering speed. The most recent methods \cite{Qian20233DGSAvatarAA, hu2024gaussianavatar, li2024animatable, zhan2025real} circumvent the rendering speed problem by using explicit 3D Gaussians and reach the best trade-off between fidelity and efficiency. 

3DGS-Avatar~\cite{Qian20233DGSAvatarAA} first proposed to use 3DGS for generating photo-realistic full-body human avatars. It unlocked real-time inference speed of 50 FPS with animated clothes reconstructed by a deformation network. To enable more realistic pose-dependent appearance, GaussianAvatar~\cite{hu2024gaussianavatar} proposed to use a Convolutional Neural Network Pose Encoder in the space of parametric SMPL~\cite{loper2023smpl} body to predict Gaussian orientation based on the human pose. AnimatableGaussians~\cite{li2024animatable} leverage modern StyleUNet \cite{wang2023styleavatar} architecture conditioned on a position map to predict pose-dependent avatar appearance. In contrast with using CNN architectures, Mmlphuman~\cite{zhan2025real} leverages spatially distributed MLPs located at different positions on the human body and achieves fine detail rendering with 150 FPS. In their approach, the Gaussian parameters are produced by interpolating from the outputs of its nearby MLPs. Alternatively to Gaussian-based approaches, MeshAvatar~\cite{chen2024meshavatar} proposes utilizing explicit triangular meshes for compatibility with the traditional graphics pipeline and incorporating physics-based rendering.


\textbf{Level of detail.} The level of detail methods can be grouped by how they define the single scene part for detail adjustment. One family of approaches determines LOD areas as hierarchical voxels. This way, Octree-GS~\cite{Ren2024OctreeGSTC} groups Gaussians in octree-structured grids and during rendering invokes corresponding detail at each tree level. Methods for large outdoor scenes \cite{Kerbl2024AH3, Kulhnek2025LODGELL, Liu2024CityGaussianRH} also tend to subdivide the scene into hierarchical chunks to efficiently render it block by block. Alternatively to square areas, \cite{Yang2025Virtualized3G} proposed to use cluster-based grouping of Gaussians with dynamic selection of their subsets to accelerate rendering. In extreme cases, Gaussians can be organized in independent levels and reconstruct the entire scene with varying detail as proposed in the FLoD~\cite{Seo2024FLoDIF}.

Another way is to not group Gaussians but to set up sparse anchors and generate Gaussians dynamically around each of them. This technique, known as neural Gaussians, is utilized by Scaffold-GS \cite{Lu2023ScaffoldGSS3} to spawn Gaussians from each anchor using an MLP. ContextGS \cite{Wang2024ContextGSC3} further develops this idea by proposing to reduce the spatial
redundancy among anchors using an autoregressive model for predicting finer levels based on coarser levels. Hash-GS~\cite{Xie2025HashGSA3} leverages high-dimensional features to parameterize primitive properties assigned to anchors via hash tables, which accelerates rendering.

While the task of LOD is well developed for static 3DGS scenes, for animatable human avatars it remains underrepresented in the literature. The existing approaches concentrate on accelerating rendering speed rather than increasing fidelity. LoDAvatar~\cite{dongye2024lodavatar} proposes to use hierarchical embedding attached to the mesh for LOD control. In contrast with our method, it generates an avatar from a 3D mesh and textures rather than directly from multi-view images. CrowdSplat~\cite{sun2024crowdsplat} generates several copies of an avatar with a different number of Gaussians to render a crowd of people efficiently. In contrast with our method, it relies on monocular input video, which limits the realism of the produced avatars and is not intended to work with close-up views. 

In this work, we target the task of highly detailed rendering of close-up views while maintaining efficiency for distant views. In contrast to previous work, this paper proposes a frequency-decomposition-based approach to determining the detail adjustment area within textured surfels~\cite{svitov2024billboard}. Namely, we propose changing billboard appearance by adjusting the texture detail for each billboard independently by mixing its high and low frequency representations.

\begin{figure}[t]
  \centering
  \includegraphics[width=12.0cm]{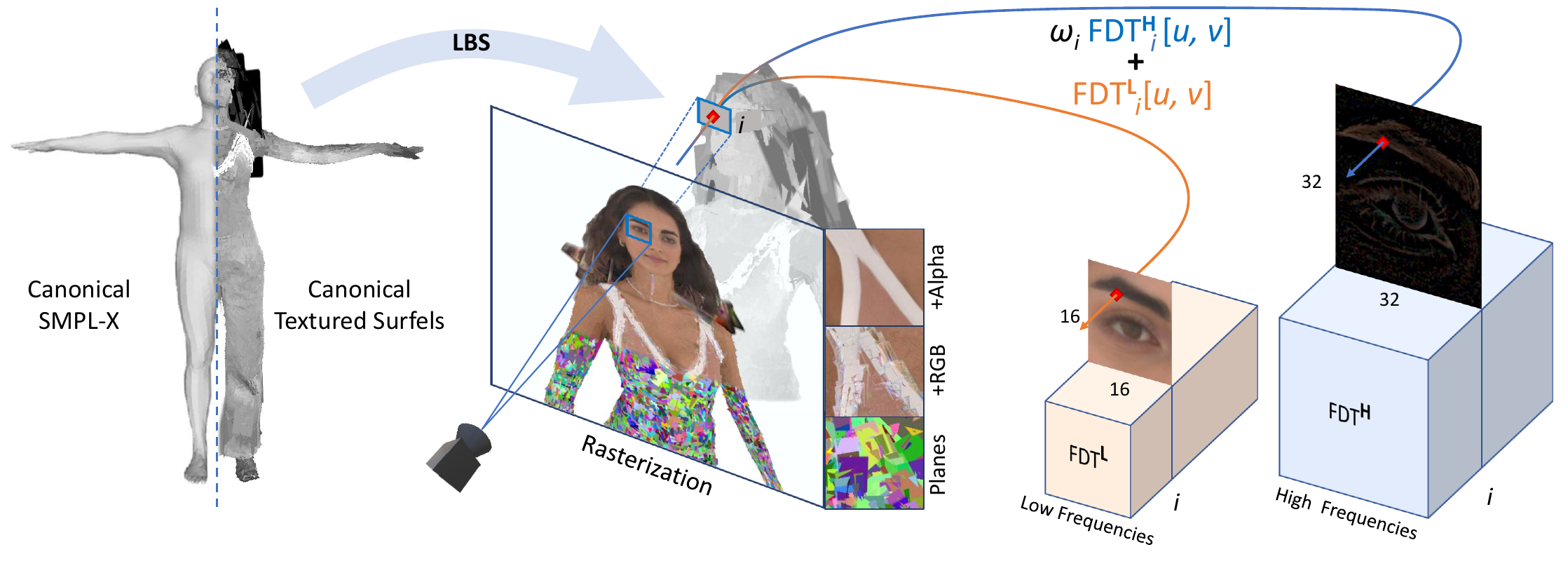}
   \caption{\textbf{Method description}. We learn the surfels' position and orientation in the canonical space and transform them to the target pose using Linear Blend Skinning (LBS). Each surfel $i$ has two learnable Frequency-Decomposed Texture (FDT), that store four channels RGB+$\alpha$ information for low $\textrm{FDT}^L$ and high $\textrm{FDT}^H$ frequency details. We store them as two tensors with corresponding textures and sample the values for surfel $i$ in $[u, v]$ coordinates. The final color of the surfel is calculated as a weighted sum with a view-dependent coefficient $\omega$. Coefficient $\omega$ controls the amount of high-frequency details based on the camera distance.}
   \label{fig:main_scheme}
\end{figure}

\section{Method}
\label{sec:method}

\subsection{Preliminaries: Billboard Splatting}
Billboard splatting was proposed in \cite{svitov2024billboard} as an alternative to 3DGS~\cite{kerbl20233d} and 2DGS~\cite{huang20242d}, and utilizes textured surfels (billboards) for 3D scene representation. In contrast with other texture-based primitives~\cite{weiss2024gaussian, rong2025gstex, song2024hdgs, xu2024supergaussians}, billboards allow both RGB and opacity texture training and enable real-time inference speed.  Billboards follow the 2DGS parametrization for location and orientation in space and additionally store learnable RGB and alpha textures. The parametrization of the billboards is as follows: $\{\mu_i, s_i, r_i, \textrm{SH}_i, T_i^\textrm{RGB}, T_i^\alpha\}$, which corresponds to the  position $\mu_i \in \mathbb{R}^3$, 2D scale $s_i\in \mathbb{R}^2$, quaternion rotation $r_i \in \mathbb{R}^4$, and spherical harmonics $\textrm{SH}_i \in \mathbb{R}^{4 \times 3}$ of the primitive. The values $\{T_i^\textrm{RGB}, T_i^\alpha\}$ correspond to learnable textures of size $S^T \times S^T$ texels for color and opacity.

During rasterization, BBSplat uses the explicit ray-splat intersection algorithm~\cite{sigg2006gpu} to find the corresponding billboard coordinate $(u, v)$ for a given screen coordinate $(x, y)$. The resulting $(u, v) \in [-1, 1]$ coordinates then rescale for the predefined texture size $S^T$ and sample values from $T_i^\textrm{RGB}$ and $T_i^\alpha$.

\subsection{Frequency-Decomposed Textures}
In this work, we propose a method for generating animatable full-body human avatars that generalize to different camera distances. We represent avatars with billboards~\cite{svitov2024billboard}, where each billboard covers a continuous area along the body surface and can be used for natural grouping of information. We estimate the proportion of frequencies' mixture for each primitive by its screen-space projection size $r$, defined as the radius of the circumscribed circle around the billboard's projection on the screen.

Billboards~\cite{svitov2024billboard} have one major drawback: the texture size $S^T$ must be selected before training. Selecting a texture that is too big or too small will affect rendering quality and training convergence. We address this limitation by proposing to use a combination of two sets of textures with different resolutions and purposes in contrast with a single set of $\{T_i^\textrm{RGB}, T_i^\alpha\}$ for the $i$-th surfel. As shown in \Cref{fig:main_scheme} we introduce two specialized set of textures $\textrm{FDT}^L \in \mathbb{R}^{4 \times 16 \times 16 \times N}$ for low-frequency and $\textrm{FDT}^H \in \mathbb{R}^{4 \times 32 \times 32 \times N}$ for high-frequency. Where each texture $T \in \{\textrm{FDT}^L, \textrm{FDT}^H\}$ contains three RGB channels $T^\textrm{RGB}$ and one opacity channel $T^\alpha$. Unlike conventional mip-mapping, our frequency levels are independently optimized rather than downsampled from a shared texture: \(FDT^L\) represents the base appearance, while \(FDT^H\) learns an additive high-frequency residual.

To sample texture in $(u, v)$ position we calculate final mixed texture $T_i[u, v] = [T_i^\textrm{RGB}[u, v], T_i^\alpha[u, v]]$ as a weighted sum of frequencies from $\textrm{FDT}_i^L$ and $\textrm{FDT}_i^H$ textures with coefficient $\omega_i$ to adjust high-frequency visibility: 

\begin{equation}
    T_i[u, v] = \textrm{FDT}_i^L[u, v] + \omega_i \textrm{FDT}_i^H[u, v].
    \label{eq:sampling}
\end{equation}

Coefficient $\omega_i \in [0, 1]$ controls how much high-frequency details we use for each billboard depending on camera position. In practice, we found it more efficient to calculate $\omega_i$ based on the billboard's screen-space projection radius $r_i$. This way, we also limit the detail of small billboards, reducing rendering artifacts. 
The equation to calculate $\omega_i$ based on $r_i$ is as follows:

\begin{equation}
    \omega_i = min(max(r_i / S_H, 0), 1),
    \label{eq:omega}
\end{equation}

where $S_H = 32$ is the $\textrm{FDT}^H$ texture size as we want to achieve full $\textrm{FDT}^H$ visibility ($\omega_i=1$) when single pixel corresponds to the single texel. Finally, we add an RGB value calculated from $\textrm{SH}_i$ to the mixed texture $T_i$ and use this texture to sample the color for each $[u, v]$ billboard point. Intuitively, instead of learning all the information in a single texture, we gradually inject high-frequency information when we close-up. Separating low and high frequencies allows us to balance scene detalization and improve convergence.

To control the avatar animation, we follow the pipeline proposed in Mmlphuman~\cite{zhan2025real} and learn billboard positions and orientations in canonical space with pose-dependent offsets $\{\delta\textbf{x}, \delta\textbf{r}, \delta\textbf{s}, \delta\textbf{c}\}$ for position, rotation, scale and opacity predicted by spatially distributed MLPs. To transform billboards from canonical space (\cref{fig:main_scheme} (left)) to the target pose, we use linear blend skinning (LBS) and transform each billboard according to the weighted SMPL-X joints' orientations. In contrast with Mmlphuman, we use magnitude fewer primitives (20K vs 200K) to represent an avatar, because each billboard contains more information than a Gaussian primitive. We selected the number of billboards to result in the same rendering speed as Gaussian-based approaches. 

\begin{figure}[t]
  \centering
  \includegraphics[width=8.5cm]{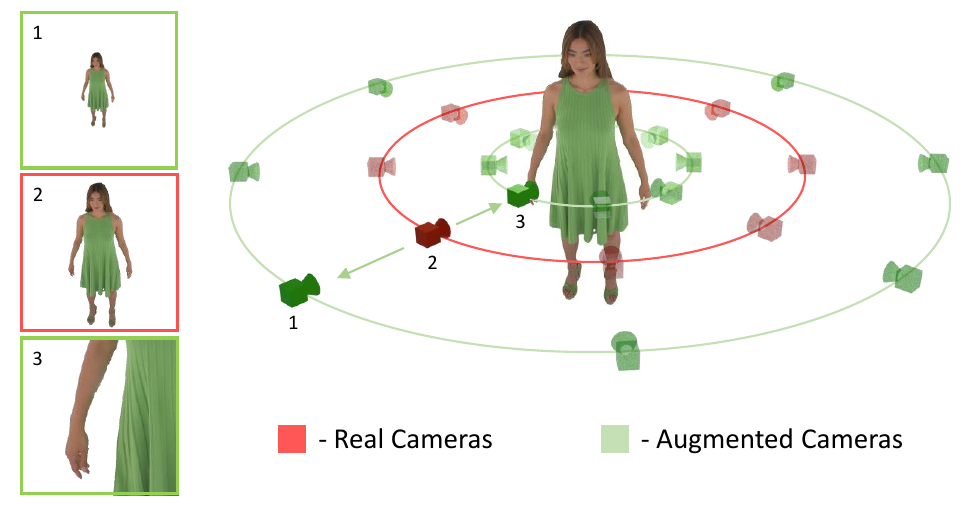}
   \caption{\textbf{Cameras augmentation}. We augment camera positions by cropping and padding dataset images to produce closer and farther views. We also modified the camera matrices as described in \Cref{sec:method_cameras} to ensure that rendering will match the modified dataset images.}
   \label{fig:cameras}
\end{figure}

\subsection{Cameras Augmentation}
\label{sec:method_cameras}
We augment cameras as shown in \Cref{fig:cameras} to ensure that avatars can be rendered from diverse camera distances. To produce far and close cameras, we pad and crop the input images accordingly. The dataset we used provides images at 4K resolution, and we noticed that it can be used to generate extremely close-up photos of the person by selecting a portion of the image, or to generate far-camera views by padding the image. We adjust camera intrinsic parameters accordingly to $W \times H$ image modification with scale factor $\mathfrak{s} = [0.25, 2.0]$ and $\Delta_x$, $\Delta_y$ offsets: 

\begin{equation}
K = \begin{bmatrix}
f_x / \mathfrak{s} & 0 & (c_x - \Delta_x)/\mathfrak{s} \\
0 & f_y / \mathfrak{s} & (c_y - \Delta_y)/\mathfrak{s}\\
0 & 0  & 1
\end{bmatrix}
\end{equation}

\begin{equation}
\Delta_x = (W - \mathfrak{s} W) / 2 \quad  \Delta_y = (H - \mathfrak{s} H) / 2
\end{equation}

Such augmentations are essential to train the method to work correctly on extreme far and close views. The proposed frequency-decomposed representation fully leverages the provided data diversity, whereas previous works struggle to converge or produce significant artifacts. Notice that we use the same number of images for training our and competing methods. We provide more details on training data in the Experiments section.

\subsection{Training}
\label{sec:method_train}
First, we initialize billboards on the SMPL-X's surface and orient them along the mesh using normal vectors. To further improve the initial positions of the billboards, we froze all textures for the first 5K iterations to adjust their initial position and learn general colors encoded with spherical harmonics (SH). We then train textures along with billboard position until 380K iterations and freeze them for the last 20K iterations to improve view-dependent SH coefficients.

During training, we want to explicitly separate low-frequency and high-frequency information between corresponding $\textrm{FDT}^L$ and $\textrm{FDT}^H$. To this end, we initialize $\textrm{FDT}^H$ with zeros and use the hyperbolic tangent activation function for both RGB and opacity channels to enforce the model to use them as offsets to the information represented by $\textrm{FDT}^L$. At the same time, we initialize the opacity of $\textrm{FDT}^L$ with a Gaussian distribution to facilitate convergence during early training steps and encourage $\textrm{FDT}^L$ to store most of the color information.

For the loss function, we utilize a combination of pixel loss, multi-scale structure similarity (MS\_SSIM), and LPIPS~\cite{zhang2018unreasonable} loss:

\begin{equation}
    \mathcal{L} = \mathcal{L}_{pixel} + \lambda_{ms\_ssim} \mathcal{L}_{ms\_ssim} + \lambda_{lpips} \mathcal{L}_{lpips},
\end{equation}

where we use $L_1$ loss as $L_{pixel}$ before 7K iterations to accelerate convergency, and $MSE$ after to increase fidelity of results. We also employed several regularization functions to ensure the avatar's surface smoothness: Scale regularization and position smoothness loss proposed in \cite{zhan2025real}, depth and normal regularizations proposed in \cite{huang20242d}. First two functions push billboards to have similar non-rigid $\delta\textbf{x}$ offsets for nearby $i, j$ and small scale $s_i$ size:

\begin{equation}
    \mathcal{L}_{ctrl} = \sum_{i, j} || \delta\textbf{x}_i - \delta\textbf{x}_j ||_2, 
    \quad
    \mathcal{L}_{scale} = \sum_{i=0}^N max(0.01, s_i)
\end{equation}

Depth and normal regularizations were first introduced in \cite{huang20242d}, and adopted for billboards in \cite{svitov2024billboard}. The normal regularization $\mathcal{L}_{norm}$ requires consistency between real normal vectors of billboards and normals calculated by the scene depth map. Depth regularization $\mathcal{L}_{depth}$ reduces discrepancy in depth between billboards along the rasterization ray. The final regularization term is a combination of the described losses:

\begin{equation}
    \mathcal{L}_{reg} = \lambda_c \mathcal{L}_{ctrl} + \lambda_s \mathcal{L}_{scale} + \lambda_n \mathcal{L}_{norm} + \lambda_d \mathcal{L}_{depth}
\end{equation}

\section{Experiments}
\label{sec:experiments}

We compared our method against three state-of-the-art full-body avatar generation approaches: Mmlphuman~\cite{zhan2025real}, AnimatableGaussians~\cite{li2024animatable} and MeshAvatar~\cite{chen2024meshavatar}. We evaluated our approach on the challenging ActorsHQ~\cite{isik2023humanrf} and THuman4.0~\cite{zheng2022structured} datasets, reporting standard objective metrics along with qualitative results. Further in this section, we discuss our choice of baselines and testing data, and provide a discussion on the results.

\begin{table}[tb]
\caption{\textbf{Camera augmentation effect.} We report metrics with and without camera distance augmentation. Baselines show metric degradation or do not converge (DNC) when trained with camera augmentation.}
\centering
\begin{adjustbox}{width=0.7\textwidth}
\begin{tabular}{l|ll|ll|l}
                     & \multicolumn{2}{c|}{w/ camera augs.}    & \multicolumn{2}{c|}{w/o camera augs.}    \\ \hline
Method                & LPIPS $\downarrow$ & FID $\downarrow$ & LPIPS $\downarrow$ & FID $\downarrow$ & FPS $\uparrow$ \\ \hline
AnimatableGaussians~\cite{li2024animatable}   & DNC & DNC & 0.3109 & 49.52 & 15 \\
Mmlphuman~\cite{zhan2025real}  & 0.3417 & 93.27  & 0.3004 & 57.16 & 251 \\
\textbf{Ours}  & \cellcolor{red!25}0.2231 & \cellcolor{red!25}38.49  & \cellcolor{red!25}0.2278 & \cellcolor{red!25}47.08 & 244
\end{tabular}
\end{adjustbox}
\label{tab:aug_ablation}
\end{table}

\begin{figure}[t]
  \centering
  \includegraphics[width=9.0cm]{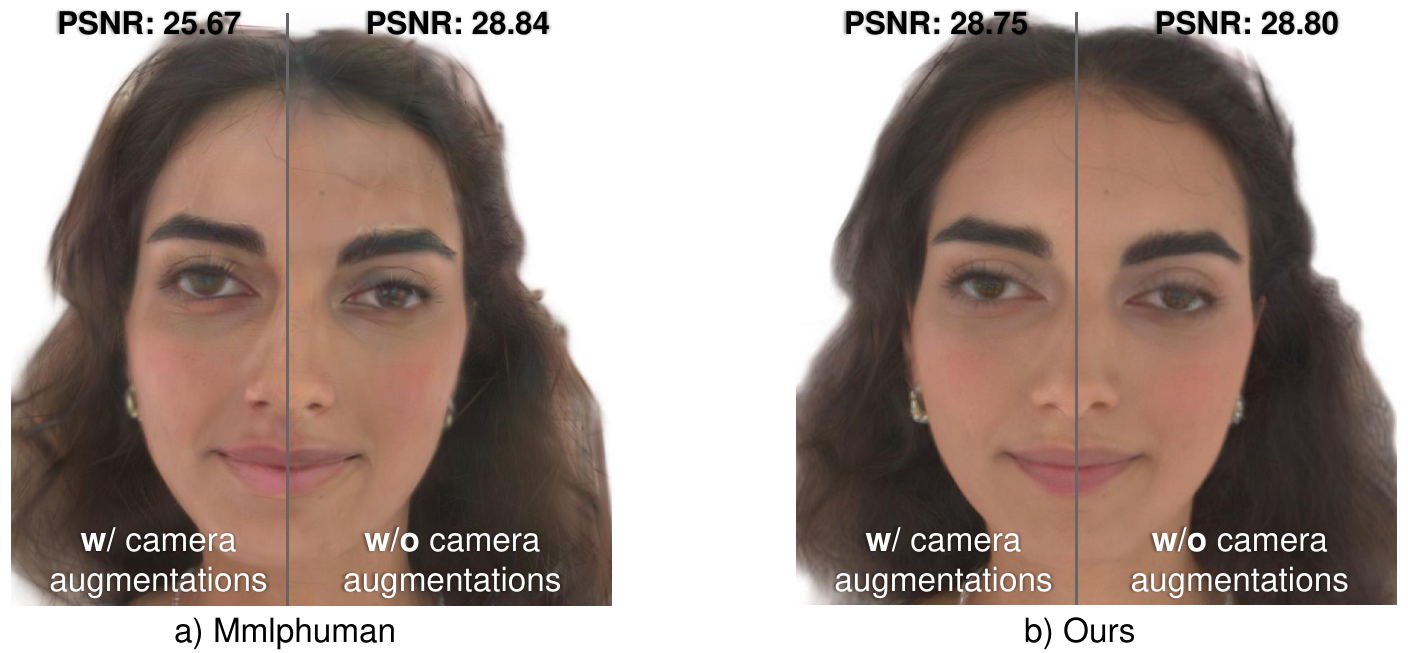}
   \caption{\textbf{Camera augmentation effect}. a) Camera augmentation described in \Cref{sec:method_cameras} negatively affects the convergence of the Gaussian-based Mmlphuman~\cite{zhan2025real} method, resulting in rendering artifacts and significant objective metrics reduction. b) For our approach, camera augmentation leads to sharper results but can cause a slight pixel-level metrics reduction due to its sensitivity to alignment with ground truth.}
   \label{fig:convergence}
\end{figure}

\begin{figure}[!t]
  \centering
  \includegraphics[width=12.0cm]{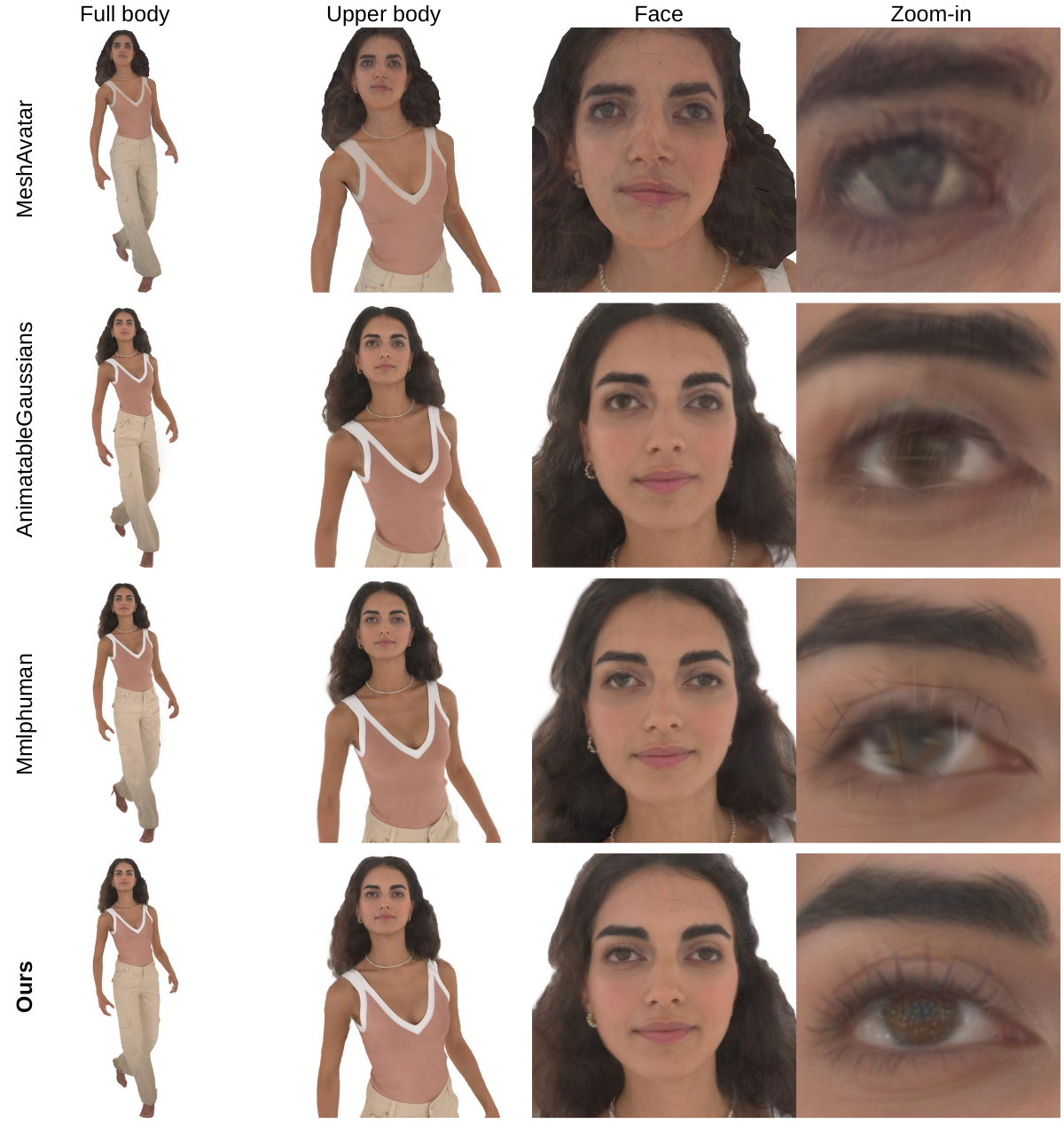}
   \caption{\textbf{Qualitative comparison for varying cameras}. Comparison of rendering quality in novel poses for novel-view synthesis across different camera distances.}
   \label{fig:zooms}
\end{figure}

\subsection{Implementation}
\label{sec:experiments_implement}
In our loss function, we set $\lambda_{SSIM} = 0.2$ following the standard 3DGS training procedure, and $\lambda_{LPIPS} = 0.1$ as in the usual avatar training pipeline. For regularization term we have chose following hyper-parameters $\lambda_c=0.1$, and $\lambda_s=0.1$ to restrict billboard non-rigid transformation. And $\lambda_n = 0.05$, $\lambda_d=100.0$ to ensure alignment of billboards along the avatar surface. We train our avatars with Adam \cite{kingma2014adam} optimizer for 400k iterations with batch size 1. For more details on training, please refer to the Supp. Mat..

We use $16 \times 16$ textures for $\textrm{FDT}^L$, and $32 \times 32$ for $\textrm{FDT}^H$ respectively. We sample from textures with different resolutions by finding the ray-plane intersection point in $(u, v)$ coordinates and reproject them to texture coordinates by multiplying by the corresponding texture size. 

\begin{table}[]

\caption{\textbf{Quantitative metrics for ActorsHQ~\cite{isik2023humanrf} dataset.} Quantitative comparison with the state-of-the-art methods for zoom-in and zoom-out camera positions. We provide a comparison with both non-real-time and real-time methods.}
\centering
\begin{adjustbox}{width=1\textwidth}
\begin{tabular}{l|lllll|lllll}
                     & \multicolumn{5}{c|}{Zoom-in}    & \multicolumn{5}{c}{Zoom-out}    \\ \hline
Method               & PSNR $\uparrow$ & SSIM $\uparrow$ & LPIPS $\downarrow$ & FID $\downarrow$ & FPS $\uparrow$ & PSNR $\uparrow$ & SSIM $\uparrow$ & LPIPS $\downarrow$ & FID $\downarrow$ & FPS $\uparrow$ \\ \hline
MeshAvatar~\cite{chen2024meshavatar}           & 24.16 & 0.7161 & 0.3191 & 66.68 & 11 & 33.10 & 0.9822 & 0.0872 & 36.07 & 27 \\
AnimatableGaussians~\cite{li2024animatable}  & \cellcolor{red!25}28.53 & \cellcolor{red!25}0.7371 & 0.3109 & \cellcolor{orange!25}49.52 & 16 & \cellcolor{orange!25}33.90 & \cellcolor{orange!25}0.9866 & \cellcolor{orange!25}0.0583 & \cellcolor{orange!25}23.69 & 15 \\ \hline\hline
Mmlphuman~\cite{zhan2025real}  & 27.64 & 0.7304 & \cellcolor{orange!25}0.3004 & 57.16 & \cellcolor{red!25}279 & 33.62 & 0.9855 & 0.0678 & 31.16 & 232 \\
CloseUpAvatar (\textbf{Ours}) & \cellcolor{orange!25}27.69 & \cellcolor{orange!25}0.7350 & \cellcolor{red!25}0.2231 & \cellcolor{red!25}38.49 & \cellcolor{orange!25}244 & \cellcolor{red!25}36.43 & \cellcolor{red!25}0.9904 & \cellcolor{red!25}0.0547 & \cellcolor{red!25}22.09 & \cellcolor{red!25}350   
\end{tabular}
\end{adjustbox}
\label{tab:zoom_metrics}
\end{table}

\begin{figure}[t]
  \centering
  \includegraphics[width=12.0cm]{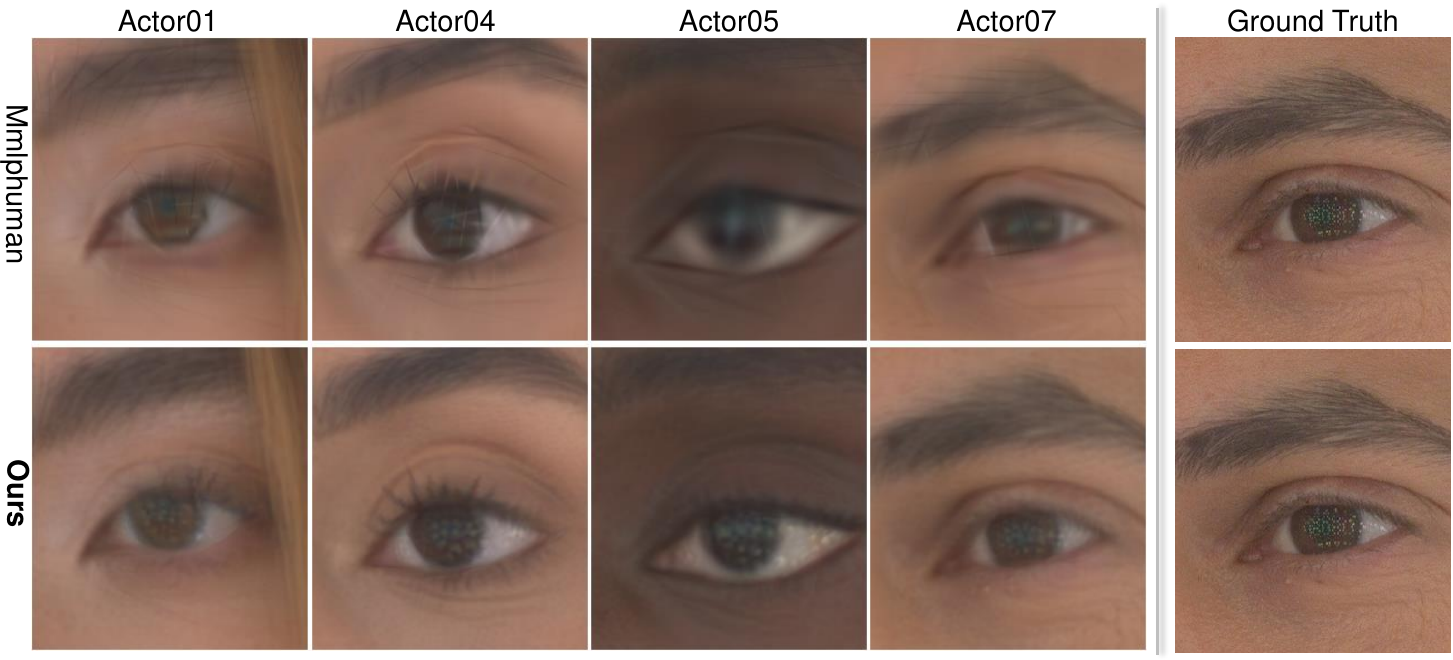}
   \caption{\textbf{Zoom-in results across different subjects}. We provide zoom-in visualizations across multiple identities from the ActorsHQ~\cite{isik2023humanrf} dataset.}
   \label{fig:other_zooms}
\end{figure}

\subsection{Results \& Evaluation}
\textbf{Baselines}.  As a baseline for the proposed approach, we selected several state-of-the-art methods for full-body avatar generation from multi-view data. Namely, we selected: Mmlphuman~\cite{zhan2025real} and AnimatableGaussians~\cite{li2024animatable} as they provide the best metrics on open datasets. Both methods use the 3DGS framework to represent avatars, and to provide a more diverse set of baselines, we also compare against the mesh-based MeshAvatar~\cite{chen2024meshavatar}. Similar to our method, MeshAvatar uses an explicit textured representation for rendering, and we demonstrated that our texture-based approach produces more realistic-looking results.


\textbf{Data}. For our experiments, we selected an open dataset, ActorsHQ~\cite{isik2023humanrf}, which provides multi-view videos of eight actors in unconstrained poses collected with 12MP Ximea cameras. The high resolution of the images in the datasets allows us to generate additional close-up view cameras by cropping images and adjusting camera matrices, as described in \Cref{sec:method_cameras}.

To ensure compatibility with evaluation protocols utilized in previous methods, we used the same cameras and frames subset as in Mmlphuman experiments. To fully utilize data capabilities, we add a few additional face-targeted cameras in all baseline training pipelines. For our method, we also enabled augmentations proposed in \Cref{sec:method_cameras}, while in \Cref{fig:convergence} and \Cref{tab:aug_ablation} we demonstrate that such augmentation negatively affects convergence for the baseline methods. Namely, Mmlphuman results in color shift and additional artifacts that lead to metrics reduction, and AnimatableGaussians does not converge under such conditions.

For compatibility with previous works, we also utilized the THuman4.0~\cite{zheng2022structured} dataset to compare under novel poses at a standard camera distance. In our experiments, we followed the THuman4.0 standard benchmark's training and testing procedures to obtain a quantitative evaluation of novel poses.

\textbf{Quantitative results}. In \Cref{tab:zoom_metrics}, we report qualitative metrics for the ActorsHQ~\cite{isik2023humanrf} dataset across novel close-up and far-away camera positions. We generate zoom-in views by taking $25\%$ central part of the input image and zoom-out views by adding $\times 2$ padding to the original image. 
We divided the table into two sections for clarity based on the inference time and provide a comparison with both real-time and non-real-time methods.  

\begin{table}[]
\caption{\textbf{Training Time.} We report training time for different methods on a single NVIDIA GeForce RTX 4090.}
\centering
\begin{tabular}{l|l}
Method               & Training Time \\ \hline
MeshAvatar           &  ~ 29 h       \\
AnimatableGaussians  &  ~ 45 h       \\
Mmlphuman            &  ~ 11 h        \\
CloseUpAvatar (\textbf{Ours}) &  \cellcolor{red!25}~ 9 h            
\end{tabular}
\label{tab:training_time}
\end{table}

For quantitative evaluation, following the previous works, we report pixel-level metrics: Peak Signal-to-Noise Ratio (PSNR) and Structural Similarity Index Measure (SSIM). We also report metrics more closely aligned with human perception: Learned Perceptual Image Patch Similarity (LPIPS~\cite{zhang2018unreasonable}) and Frechet Inception Distance (FID~\cite{heusel2017gans}). 

We achieve the best value across all metrics for far-away camera views while achieving the highest inference speed. For close-up views, we reach the best perception metrics preserving real-time inference speed. While we achieve the best perception scores, we fall slightly behind AnimatableGaussians on PSNR and SSIM, as these metrics are sensitive to pixel offsets in sharper images, reflecting the known perception-distortion trade-off~\cite{blau2018perception}.
While AnimatableGaussians provides the best competitor metrics, both it and MeshAvatar are considerably slower due to large convolutional networks for modeling pose-dependent effects.


Our CloseUpAvatar reaches better objective metrics than Mmlphuman while maintaining real-time inference speed. In the zoom-out scenario, we achieve an even higher speed, which can be practical for rendering multiple people. We reach high inference speed by using a magnitude fewer primitives (20K instead of 200K for Mmlphuman). In combination with high texture convergence speed, we also achieve a faster training procedure, as shown in \Cref{tab:training_time}.

Finally, to evaluate quantitative metrics for novel poses, we use THuman4.0~\cite{zheng2022structured} benchmark following previous works ~\cite{li2024animatable, zhan2025real}. In \Cref{tab:benchmark}, we compare against the best metric competitor \cite{li2024animatable} and the real-time competitor \cite{zhan2025real}. Our CloseUpAvatar reaches the best perceptual metrics (LPIPS, FID) while achieving the fastest rendering on our NVIDIA GeForce RTX 4090 GPU. 

\begin{table}[]
\caption{\textbf{Quantitative metrics for THuman4.0~\cite{zheng2022structured} dataset.} We report metrics for the methods trained and tested using the standard THuman4.0 benchmark to evaluate an avatar under novel poses.}
\centering
\begin{tabular}{l|lllll}
Method           &  LPIPS $\downarrow$ & FID $\downarrow$ & PSNR $\uparrow$ & SSIM $\uparrow$ & FPS $\uparrow$ \\ \hline
AnimatableGaussians~\cite{li2024animatable} & 0.0515 & 29.48 & \cellcolor{red!25}28.07 & \cellcolor{red!25}0.9739 & 15 \\
Mmlphuman~\cite{zhan2025real}  & 0.0492 & 28.06 & 27.87 & 0.9736 & 287 \\ %
CloseUpAvatar (\textbf{Ours}) & \cellcolor{red!25}0.0488 & \cellcolor{red!25}27.95 & 27.66 & 0.9730 & \cellcolor{red!25}407 %
\end{tabular}
\label{tab:benchmark}
\vspace{-10pt}
\end{table}

\begin{figure}[t]
  \centering
  \includegraphics[width=12.0cm]{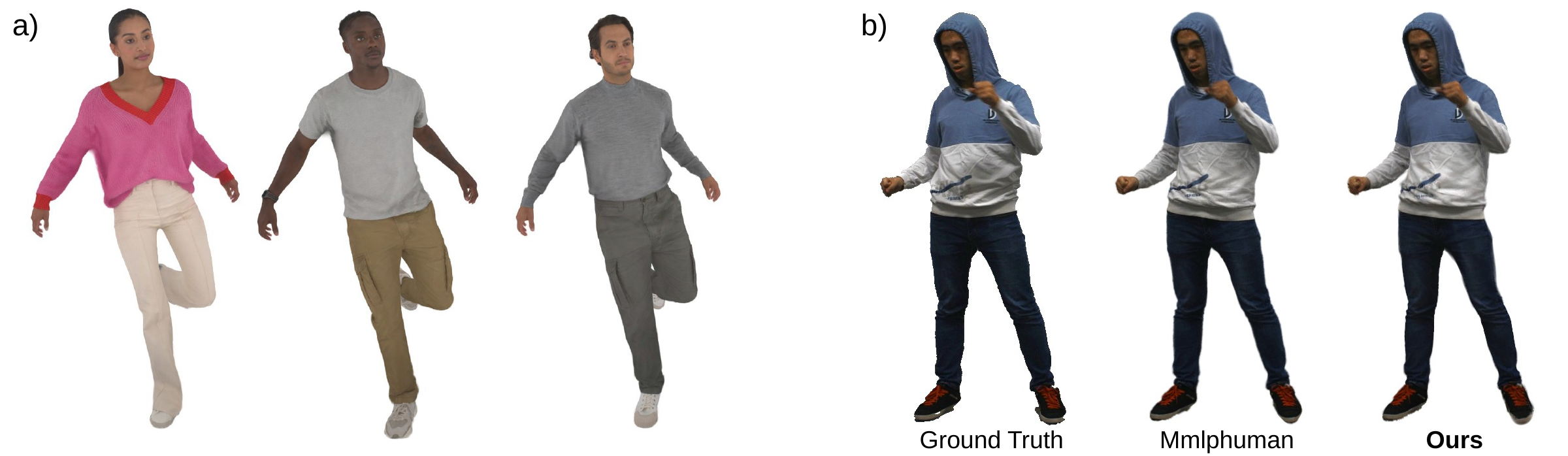}
   \caption{\textbf{Novel poses}. a) Showcase of our avatars in novel poses from the AMASS~\cite{mahmood2019amass} dataset. b) Comparison of avatars under novel poses with real-time Mmlphuman~\cite{zhan2025real} method utilizing THuman4.0~\cite{zheng2022structured} benchmark.}
   \label{fig:poses}
\end{figure}

\textbf{Qualitative results}. In \Cref{fig:zooms}, we showcase a comparison of the proposed method with the state-of-the-art for different camera distance scenarios. The avatars shown in novel poses from the AMASS \cite{mahmood2019amass} dataset, from the novel camera positions. We demonstrate more zoom-ins for different subjects in \Cref{fig:other_zooms}. We also report visualizations of our full-body avatars in novel poses in \Cref{fig:poses}. We demonstrate avatars for diverse actors from the ActorsHQ~\cite{isik2023humanrf} dataset (\cref{fig:poses}~(a)), and for Subject00 from THuman4.0~\cite{zheng2022structured} dataset (\cref{fig:poses}~(b)). Our method provides sharper results compared to Mmlphuman due to the use of trainable textures.

While previous approaches produce realistic quality for animatable full-body avatars, they tend to produce artifacts in closer views, such as overlapping or shifted Gaussians, or overblurred results. Our CloseUpAvatar maintains good quality for far and average views, with improved sharpness of edges by better fitting them with textures. For challenging close-up views, we are able to produce photo-realistic rendering, enabling more practical applications for full-body avatars in VR. In \Cref{fig:convergence}, we demonstrate that improved close-view rendering quality is achieved by using the proposed frequency-decomposed textures representation rather than a wider camera distance range.

\subsection{Ablation Study} 
\begin{figure}[t]
  \centering
  \includegraphics[width=12.0cm]{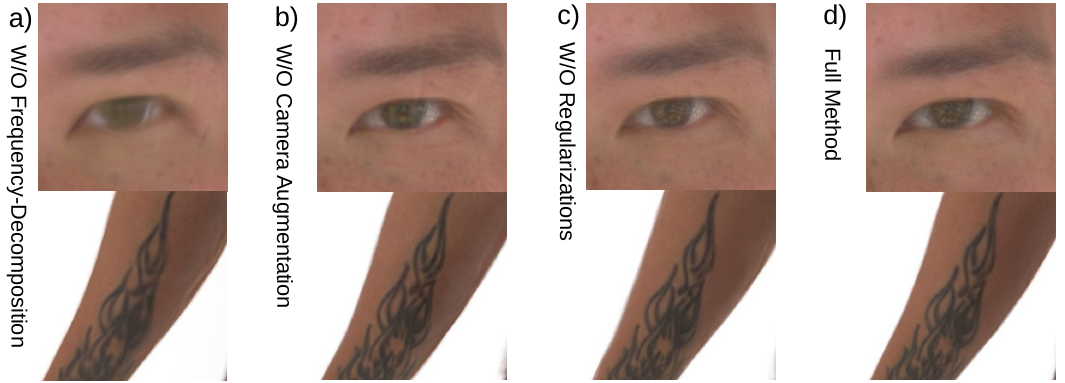}
   \caption{\textbf{Ablation study}. We demonstrate the quality of zoom-ins with ablated key components of the proposed method.}
   \label{fig:ablation}
\end{figure}

We conducted an ablation study of the key components of the proposed approach in \Cref{fig:ablation} and report corresponding objective metrics in \Cref{tab:ablation}. Namely, we ablated the use of frequency specific textures and instead used a single large $32 \times 32$ texture for each billboard (\cref{fig:ablation} (a)). The use of a standard single texture for each billboard results in significantly more blurred renders, which negatively affects LPIPS and FID score. We also demonstrated that our method can work in scenarios where cameras of different distances are not available by disabling camera augmentation (\cref{fig:ablation} (b)). It resulted in minor rendering artifacts and blurriness, which also affect perception metrics. 

Finally, in \Cref{fig:ablation} (c), we demonstrate the avatar trained without using the regularizations described in \Cref{sec:method_train}. Namely, surfels scale regularization $\mathcal{L}_{scale}$, normals consistency regularization $\mathcal{L}_{norm}$, neighboring billboards offsets similarity regularization $\mathcal{L}_{ctrl}$, and depth discrepancy regularization $\mathcal{L}_{depth}$. While the overall quality of the avatar is on par with the full method ((\cref{fig:ablation}~(d)), the disabling of regularization terms leads to a noticeable reduction in objective metrics (\cref{tab:ablation}).

Thus, we can conclude that all parts of the proposed approach are essential to get the best qualitative result and objective perception metrics. Notice that while the pixel-level SSIM metric can be improved by disabling frequency-based decomposition, this results in blurred renders and a significant drop in perception scores (LPIPS \& FID).

\begin{table}[]
\caption{\textbf{Ablation study.} We report metrics for the proposed method with ablated key components.}
\centering
\begin{tabular}{l|llll}
                        & PSNR$\uparrow$ & SSIM$\uparrow$ & LPIPS$\downarrow$ & FID$\downarrow$ \\ \hline
w/o Frequency Tex.    & 27.29 & \cellcolor{red!25}0.7375 & 0.2842 & 60.03 \\
w/o Camera Aug.         & 27.47 & 0.7350 & 0.2810 & 48.57 \\
w/o Regularizations     & 27.62 & 0.7334 & 0.2278 & 47.08 \\ \hline
Full Method             & \cellcolor{red!25}27.69 & 0.7350 & \cellcolor{red!25}0.2231 & \cellcolor{red!25}38.49
\end{tabular}
\label{tab:ablation}
\end{table}

\section{Conclusion and Limitation}
\label{sec:conclusion}
In this paper, we propose a method for photo-realistic rendering of full-body human avatars from different camera distances. The proposed approach outperforms existing methods both quantitatively and qualitatively, as we have shown in extensive experiments. We are able to render human avatars at real-time inference speed and preserve a high level of detail even for extreme close-up views.

While the proposed method successfully solves the task of detailed rendering from different camera positions, it still has a few limitations. It can be challenging for textured surfels to accurately handle non-rigid transformation due to the comparatively large size of primitives. Also, this work focused on preserving detail in rendering and overlooked questions of fine fingers and face animation. We plan to investigate this topic in our future work.

%
%
\bibliographystyle{splncs04}
\bibliography{main}

\clearpage
\setcounter{section}{0}
\renewcommand{\thesection}{\Alph{section}}

 
\let\titleold\title
\renewcommand{\title}[1]{\titleold{#1}\newcommand{\thetitle}{#1}}
\def\maketitlesupplementary
   {
   \onecolumn
   \begin{center}
   \Large
    \textbf{\thetitle}\\
    \small
    \vspace{0.5em}------------------------------ \\
    \vspace{1.0em}
   \end{center}
   }
 

\title{Supplementary materials for "Frequency-Decomposed Avatar Representation for Varying Camera Distances"}
\maketitlesupplementary

\section{Testing Procedure}
\begin{figure}[t]
  \centering
  \includegraphics[width=8.5cm]{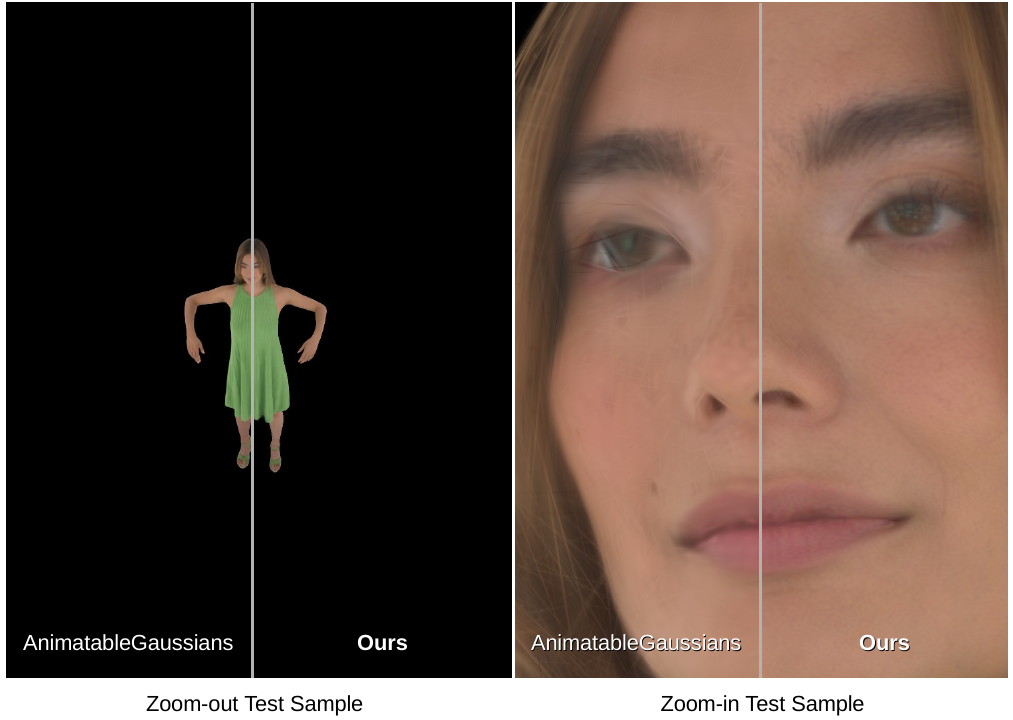}
   \caption{\textbf{Test view samples}. The image is a showcase of camera views used for quantitative evaluation in \Cref{tab:zoom_metrics}. We report samples for our and the best among competitors AnimatableGaussians methods. For better details please zoom in.}
   \label{fig:test_samples}
\end{figure}

In \Cref{tab:zoom_metrics} we report quantitative metrics for zoom-in and zoom-out views. In \Cref{fig:test_samples} we report examples of zoom-out (in the left) and zoom-in (in the right) views used in test experiments. We produce zoom-out views by $\times 2$ reduction of the dataset image, and zoom-in views by taking a 25\% crop of the face targeted camera. We specifically targeted zoom-in views for faces for each person, because humans are more perceptive to this area, and we targeted our evaluation there as well. Please note that we used a black background for testing images to explicitly zero background impact in the metrics.

We also report per-person quantitative metrics in \Cref{tab:scene_results} for both zoom-in and zoom-out scenarios. While we achieve the best LPIPS and FID perception scores in most scenarios, we fall behind in pixel-level PSNR and SSIM. This is caused by producing sharper output as shown in \Cref{fig:test_samples}. Pixel-level metrics tend to show better values for slightly blurred results, because on average, they better fit ground truth images. This problem is especially noticeable in human avatars due to small body movements, such as blinking or wrinkle movements. Therefore, we chose two convolution neural network-based (LPIPS \& FID) as they are known for better correlation with human perception. 

\section{Training Details}
In this section, we first discuss the initialization procedure for the textured surfels along the SMPL-X mesh surface. To ensure convergence, we align surfels with the mesh's normal vectors. To this end we first initialize surfels in positions $\mathbf{\mu}_i  \in \mathbb{R}^3$ along SMPL-X surface, and find three nearest surfels $\mathbf{\mu}_i^1, \mathbb{\mu}_i^2, \mathbb{\mu}_i^3 \in \mathbb{R}^3$ to each $\mathbf{\mu}_i$. Then we find the corresponding edge vectors as:
\begin{equation}
\mathbf{e}_i^1 = \mathbf{\mu}_i^1 - \mathbf{\mu}_i, \quad
\mathbf{e}_i^2 = \mathbf{\mu}_i^2 - \mathbf{\mu}_i, \quad
\mathbf{e}_i^3 = \mathbf{\mu}_i^3 - \mathbf{\mu}_i.
\end{equation}
The surfel normal is estimated as the average of the cross products of consecutive edge vectors:
\begin{equation}
\mathbf{n}_i = \frac{1}{K} \sum_{k=1}^{K} \mathbf{e}_i^k \times \mathbf{e}_i^{(k+1) \bmod K},
\end{equation}
where (K=3) in our case. After normalizing $\mathbf{n}_i$ by dividing on $||\mathbf{n}_i||$ to produce $\hat{\mathbf{n}}_i$, we assemble rotation matrix $\mathbf{R}_i =
\begin{bmatrix}
\mathbf{t}_i & \mathbf{b}_i & \hat{\mathbf{n}}_i
\end{bmatrix}.$ Here $\mathbf{t}_i$ and $\mathbf{b}_i$ are correspondingly the tangent direction and binominal direction produced by multiplying with “up” direction vector $\mathbf{u} = [0, 1, 0]^\top$:
\begin{equation}
    \mathbf{t}_i = \frac{\mathbf{u} \times \hat{\mathbf{n}}_i}{|\mathbf{u} \times \hat{\mathbf{n}}_i|} \quad \mathbf{b}_i = \frac{\hat{\mathbf{n}}_i \times \mathbf{t}_i}{|\hat{\mathbf{n}}_i \times \mathbf{t}_i|}.
\end{equation}

The resulting rotation matrix $\mathbf{R}_i$, we then transform to a quaternion representation $\mathbf{r}_i = \text{quat}(\mathbf{R}_i)$ and use it to initialize the surfels' orientation.

Beside of hyper-parameters provided in the \Cref{sec:experiments_implement} we utilized annealing of trainable parameters. For training all parameters of the avatar with Adam optimizer, we applied exponential learning rate annealing by multiplying them by $\gamma$. We set $\gamma$ parameter of annealing with $0.01^{1 / M}$, where $M$ = 400K is the number of iterations.

We disable depth $\mathcal{L}_{depth}$ and normal $\mathcal{L}_{norm}$ regularisations for the first iterations to allow the avatar to first freely optimize surfels' orientation. Then we enabled depth regularization after 3K iterations and normal regularization after 7K iterations.

\section{Results Visualization}
\begin{figure*}[t]
  \centering
  \includegraphics[width=12.0cm]{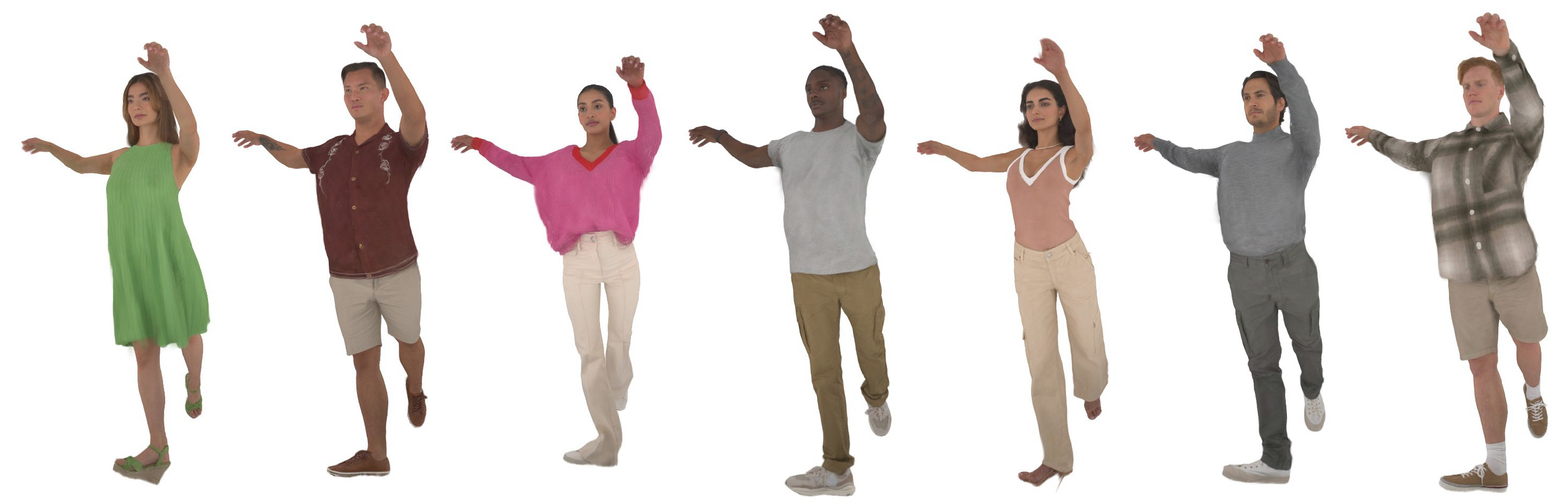}
   \caption{\textbf{Novel poses for all actors from ActorsHQ}. We show novel poses and views renders for all actors from the ActorsHQ dataset. While we are able to provide realistic close-up renders, we preserve the ability to render novel poses with high fidelity.}
   \label{fig:more_poses}
\end{figure*}

\begin{figure*}[t]
  \centering
  \includegraphics[width=12.0cm]{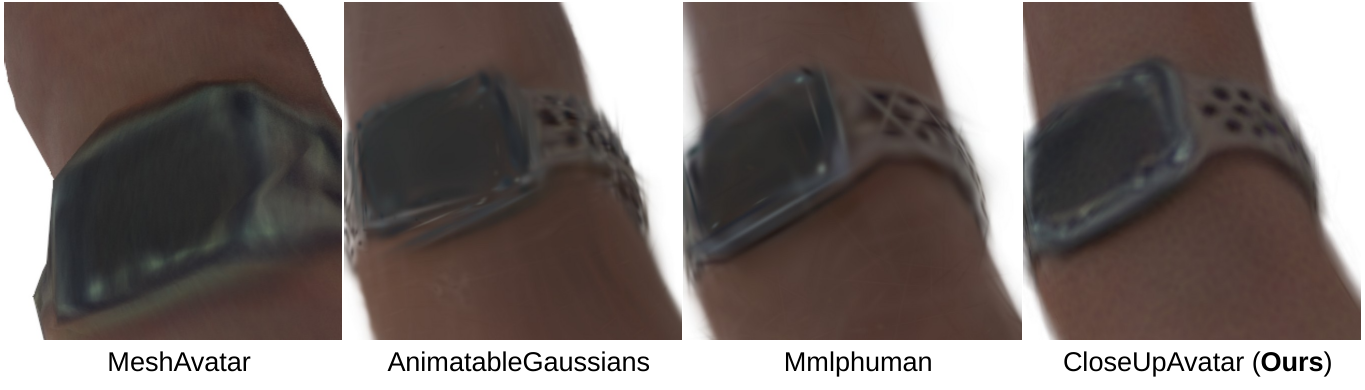}
   \caption{\textbf{Zoom-in on non-facial area}. We provide close-up renders of the hand with a watch of Actor05. Our method provides more detailed rendering (\eg circle holes on the bracelet).}
   \label{fig:hands}
\end{figure*}

In \Cref{fig:more_poses} we showcase more visualizations of avatars in novel poses, that were initially presented in \Cref{fig:poses}. We also report more zoom-in examples for non-facial areas in \Cref{fig:hands}. Our method is able to represent more details with a limited number of surfels (only 20K). 

\begin{table*}[]

\caption{\textbf{Quantitative metrics for all persons.} We report pixel-level (PSNR \& SSIM), and perception (LPIPS \& FID) metrics for each actor in the ActorsHQ dataset for zoom-in and zoom-out views. Averaged values can be found \Cref{tab:zoom_metrics}.}

\fontsize{8}{12}\selectfont
\centering
\begin{adjustbox}{width=1\textwidth}
\begin{tabular}{cc|cccc}
                                &            & MeshAvatar    & AnimatableGaussians       & Mmlphuman  & Ours  \\ 
                                &            & \footnotesize{PSNR$\uparrow$ / SSIM$\uparrow$ / LPIPS$\downarrow$ / FID$\downarrow$} & \footnotesize{PSNR$\uparrow$ / SSIM$\uparrow$ / LPIPS$\downarrow$ / FID$\downarrow$} & \footnotesize{PSNR$\uparrow$ / SSIM$\uparrow$ / LPIPS$\downarrow$ / FID$\downarrow$} & \footnotesize{PSNR$\uparrow$ / SSIM$\uparrow$ / LPIPS$\downarrow$ / FID$\downarrow$} \\ \hline
\parbox[t]{2mm}{\multirow{7}{*}{\rotatebox[origin=c]{90}{\footnotesize{Zoom-in}}}} 
                                & Actor 01 &  23.84 / 0.6928 / 0.3142 / 70.90 & \colorbox{orange!25}{27.72} / 0.7169 / 0.2892 / \colorbox{orange!25}{55.31} & \colorbox{red!25}{27.95} / \colorbox{orange!25}{0.7171} / \colorbox{orange!25}{0.2708} / 64.79 & 27.23 / \colorbox{red!25}{0.7174} / \colorbox{red!25}{0.2168} / \colorbox{red!25}{46.03} \\
                                & Actor 02 &  22.55 / 0.7112 / 0.3248 / 81.39 & \colorbox{red!25}{27.35} / \colorbox{red!25}{0.7276} / \colorbox{orange!25}{0.3133} / \colorbox{orange!25}{48.45} & 25.08 / 0.7129 / 0.3222 / 54.65 & \colorbox{orange!25}{26.64} / \colorbox{orange!25}{0.7256} / \colorbox{red!25}{0.2188} / \colorbox{red!25}{33.71} \\
                                & Actor 04 &  21.71 / 0.7531 / 0.2855 / 41.94 & \colorbox{red!25}{25.65} / \colorbox{red!25}{0.7808} / 0.2839 / 42.65 & \colorbox{orange!25}{24.60} / 0.7676 / \colorbox{orange!25}{0.2723} / \colorbox{orange!25}{36.67} & 24.33 / \colorbox{orange!25}{0.7751} / \colorbox{red!25}{0.2062} / \colorbox{red!25}{30.73} \\
                                & Actor 05 &  26.55 / 0.7099 / 0.3469 / 71.60 & \colorbox{red!25}{29.81} / \colorbox{red!25}{0.7263} / 0.3475 / \colorbox{orange!25}{66.83} & \colorbox{orange!25}{28.58} / 0.7207 / \colorbox{orange!25}{0.3364} / 92.45 & 28.35 / \colorbox{orange!25}{0.7232} / \colorbox{red!25}{0.2574} / \colorbox{red!25}{44.73} \\
                                & Actor 06 &  24.36 / 0.7213 / 0.3061 / 43.70 & \colorbox{red!25}{30.03} / \colorbox{red!25}{0.7525} / 0.2980 / 39.39 & \colorbox{orange!25}{28.84} / 0.7457 / \colorbox{orange!25}{0.2794} / \colorbox{orange!25}{36.52} & 28.75 / \colorbox{orange!25}{0.7466} / \colorbox{red!25}{0.2096} / \colorbox{red!25}{29.44} \\
                                & Actor 07 &  24.06 / 0.6955 / 0.3350 / 79.62 & \colorbox{red!25}{29.65} / \colorbox{red!25}{0.7125} / 0.3145 / \colorbox{orange!25}{53.87} & 28.48 / 0.7045 / \colorbox{orange!25}{0.3130} / 63.96 & \colorbox{orange!25}{29.12} / \colorbox{orange!25}{0.7113} / \colorbox{red!25}{0.2264} / \colorbox{red!25}{37.46} \\
                                & Actor 08 &  26.05 / 0.7291 / 0.3214 / 77.61 & \colorbox{orange!25}{29.48} / 0.7427 / 0.3299 / \colorbox{red!25}{40.14} & \colorbox{red!25}{29.93} / \colorbox{orange!25}{0.7441} / \colorbox{orange!25}{0.3085} / 51.13 & 29.42 / \colorbox{red!25}{0.7460} / \colorbox{red!25}{0.2263} / \colorbox{orange!25}{47.37} \\
                                \hline \hline
\parbox[t]{2mm}{\multirow{7}{*}{\rotatebox[origin=c]{90}{\footnotesize{Zoom-out}}}} 
                                & Actor 01 &  33.10 / 0.9824 / 0.0836 / 39.88 & \colorbox{orange!25}{34.64} / \colorbox{orange!25}{0.9878} / \colorbox{red!25}{0.0505} / \colorbox{red!25}{17.70} & 34.23 / 0.9867 / 0.0583 / 23.56 & \colorbox{red!25}{36.73} / \colorbox{red!25}{0.9906} / \colorbox{orange!25}{0.0523} / \colorbox{orange!25}{18.22} \\
                                & Actor 02 &  33.73 / 0.9813 / 0.0912 / 39.20  & \colorbox{orange!25}{34.69} / \colorbox{orange!25}{0.9858} / \colorbox{orange!25}{0.0583} / \colorbox{orange!25}{30.63} & 33.41 / 0.9825 / 0.0756 / 41.82 & \colorbox{red!25}{37.02} / \colorbox{red!25}{0.9897} / \colorbox{red!25}{0.0542} / \colorbox{red!25}{23.69} \\
                                & Actor 04 &  32.20 / 0.9840 / 0.0901 / 26.57 & 33.23 / 0.9877 / \colorbox{orange!25}{0.0637} / \colorbox{orange!25}{17.25} & \colorbox{orange!25}{33.41} / \colorbox{orange!25}{0.9878} / 0.0703 / 22.10 & \colorbox{red!25}{36.12} / \colorbox{red!25}{0.9913} / \colorbox{red!25}{0.0556} / \colorbox{red!25}{15.55} \\
                                & Actor 05 &  33.00 / 0.9828 / 0.0854 / 37.24 & \colorbox{orange!25}{34.30} / \colorbox{orange!25}{0.9862} / \colorbox{orange!25}{0.0626} / \colorbox{red!25}{23.33} & 34.22 / 0.9854 / 0.0703 / 34.95 & \colorbox{red!25}{36.51} / \colorbox{red!25}{0.9905} / \colorbox{red!25}{0.0592} / \colorbox{orange!25}{24.11} \\
                                & Actor 06 &  33.06 / 0.9858 / 0.0795 / 40.83 & 33.07 / \colorbox{orange!25}{0.9877} / \colorbox{orange!25}{0.0560} / \colorbox{orange!25}{39.36} & \colorbox{orange!25}{33.13} / 0.9876 / 0.0652 / 43.63 & \colorbox{red!25}{35.91} / \colorbox{red!25}{0.9918} / \colorbox{red!25}{0.0536} / \colorbox{red!25}{36.82} \\
                                & Actor 07 & \colorbox{orange!25}{35.04} / 0.9842 / 0.0876 / 36.65 & 34.56 / \colorbox{orange!25}{0.9865} / \colorbox{orange!25}{0.0619} / \colorbox{orange!25}{23.10} & 33.93 / 0.9845 / 0.0755 / 34.01 & \colorbox{red!25}{37.12} / \colorbox{red!25}{0.9903} / \colorbox{red!25}{0.0565} / \colorbox{red!25}{22.55} \\
                                & Actor 08 &  31.59 / 0.9748 / 0.0929 / 32.16 & 32.78 / \colorbox{orange!25}{0.9847} / \colorbox{orange!25}{0.0549} / \colorbox{orange!25}{14.51} & \colorbox{orange!25}{33.01} / 0.9842 / 0.0593 / 18.05 & \colorbox{red!25}{35.63} / \colorbox{red!25}{0.9885} / \colorbox{red!25}{0.0519} / \colorbox{red!25}{13.73} \\
                                \hline
                                
\end{tabular}
\end{adjustbox}
\label{tab:scene_results}
\end{table*}

\end{document}